\documentclass[letterpaper, 10 pt, conference]{ieeeconf}  

\IEEEoverridecommandlockouts                              

\usepackage{graphics} 
\usepackage{amssymb}  
\usepackage{algorithm}
\usepackage{algpseudocode}
\usepackage{mathrsfs}

\usepackage{amsmath}
\usepackage{hyperref}
\usepackage{mathtools}
\usepackage{bm}
\usepackage{amssymb}
\usepackage{graphicx}
\usepackage{caption}
\usepackage{multicol}
\usepackage{multirow}
\usepackage{algorithm}
\usepackage[makeroom]{cancel}
\usepackage{algpseudocode}
\usepackage{subcaption}
\usepackage{wasysym}
\usepackage{tikz}
\usepackage{siunitx}

\newcommand{\E}{\mathbb{E}}

\title{\LARGE \bf
Robotic Arm Control and Task Training\\ through Deep Reinforcement Learning
}

\author{Andrea Franceschetti$^{1}$, Elisa Tosello$^{1}$, Nicola Castaman$^{1}$, and Stefano Ghidoni$^{1}$
\thanks{$^{1}$A. Franceschetti, E. Tosello, N. Castaman and S. Ghidoni are with the Intelligent Autonomous Systems Lab (IAS-Lab), Deptartment of Information Engineering, University of Padova, Via Gradenigo 6/B, 35131 Padova, Italy
        {\tt\small [andrea.franceschetti@studenti.unipd.it, toselloe@dei.unipd.it, castaman@dei.unipd.it, ghidoni@dei.unipd.it]}}%
}

\begin{document}

\maketitle
\thispagestyle{empty}
\pagestyle{empty}


\begin{abstract}\label{abstract}
This  paper  proposes  a  detailed  and  extensive  comparison  of  the  Trust  Region  Policy  Optimization and  Deep  Q-Network  with  Normalized  Advantage  Functions  with  respect  to  other state  of  the art  algorithms,  namely  Deep  Deterministic  Policy  Gradient and Vanilla Policy  Gradient. Comparisons demonstrate that the former have better performances then the latter when asking robotic arms to accomplish manipulation tasks such as reaching a random target pose and pick \& placing an object. Both simulated and real-world experiments are provided. Simulation lets us show the procedures that we adopted to precisely estimate the algorithms hyper-parameters and to correctly design good policies. Real-world experiments let show that our polices, if correctly trained on simulation, can be transferred and executed in a real environment with almost no changes. 
 
\end{abstract}

\section{INTRODUCTION}\label{introduction}
Modern robots operating in real environments should be able to cope with dynamic workspaces. They should autonomously and flexibly adapt to new tasks, new motions, environment changes, and disturbances. These requirements generated novel challenges in the area of Robot Control. It is no longer sufficient to implement control algorithms that are robust to noise; they should also become independent from the assigned task, the planned motion, and the accuracy of the dynamic model of the system to be controlled. They should easily adapt to different devices and working conditions by overcoming the need of complex parameters identification and/or system re-modeling.

Reinforcement Learning (RL) has been commonly adopted for this purpose. However, the high number of degrees-of-freedom of modern robots leads to large dimensional state spaces, which are difficult to be learned: example demonstrations must often be provided to initialize the policy and mitigate safety concerns during training. Moreover, when performing dimensionality reduction, not all the dimensions can be fully modelled: an appropriate representation for the policy or value function must be provided in order to achieve training times that are practical for physical hardwares. 

The conjunction of parallel computing and Embedded Deep Neural Networks (DNN) extended RL to continuous control applications. Parallel computing provides concurrency, particularly performing simultaneously multiple actions at the same time. DNN overcomes the need for infinite memory for storing experiences: it approximates non-linear multidimensional functions by parametrizing agents (i.e., robots) experiences through the network's finite weights. The notion of Deep Reinforcement Learning (DRL) results.

This paper proposes a detailed and extensive comparison of the Trust  Region  Policy  Optimization (TRPO)~\cite{schulman2015trust} and Deep Q-Network with Normalized Advantage Functions (DQN-NAF)~\cite{gu2017deep} with respect to other state of the art algorithms, namely Deep Deterministic Policy Gradient (DDPG)~\cite{silver2014deterministic} and Vanilla Policy Gradient (VPG)~\cite{williams1992simple}. Both simulated and real-world experiments are provided. They let to finely describe the hyper-parameters selection and tuning procedures as well as demonstrate the robustness and adaptability of TRPO and DQN-NAF while performing manipulation tasks such as reaching a random position target and pick \& placing an object. Such algorithms are able to learn new manipulation policies from scratch, without user demonstrations and without the need of a task-specific domain knowledge. Moreover, their model-freedom guarantees good performances even in case of changes in the dynamic and geometric models of the robot (e.g., link lengths, masses, and inertia). 

The rest of the paper is organized as follows. Section~\ref{state_of_the_art} describes existing DRL algorithms. In Section~\ref{system} the essential notation is introduced together with the foundations of RTPS and DQN-NAF. Section~\ref{simulated_experiments} describes the simulated experiments: a detailed description of the implemented simulated robot model is depicted, together with the tasks we ask it to perform. Simulation lets us deduce and prove the correctness of our system design as well as show the steps to follow for a powerful hyper-parameters estimation. Section ~\ref{real_experiments} transfers our policies to a real setup. 
Finally, Section~\ref{conclusions} contains conclusions and future works.


\section{STATE OF THE ART}\label{state_of_the_art}
During the years, successful applications of NNs for robotics systems have been implemented. Among others, fuzzy neural networks and explanation-based neural networks have allowed robots to learn basic navigation tasks. Multi-Layer Perceptrons (MLPs) were adopted to learn various tasks of the RoboCup soccer challenge, e.g., defenses, interception, kicking, dribbling and penalty shots. With respect to Robot Control, neural oscillators with sensor feedback have been used to learn rhythmic movements where open and closed-loop information were combined, such as gaits for a two legged robot. Focusing on model-free DRL, \cite{gu2016continuous} and \cite{gu2017deep} make a robotic arm learn to open a door from scratch by using DQN-NAF. \cite{andrychowicz2017hindsight}, instead, uses Hindsight experience Replay (HER) to train several tasks by assigning sparse and binary rewards. Such tasks include the Pick\&Place of an object and the pushing of a cube. Recently, the scientific community is achieving notable results by combining model-free and model-based DRL with Guided Policy Search (GPS) \cite{levine2016end, yahya2017collective,chebotar2017combining}. This combination guarantees good performances on various real-world manipulation tasks requiring localization, visual tracking and complex contact dynamics tasks. \cite{levine2016learning}, instead, manages to train from single view image streams a neural network able to predict the probability of successful grasps, learning thus a hand-eye coordination for grasping. Interesting related works on visual DRL for robotics are also \cite{lee2017learning, mahler2017dex, perez2017c,zhang2015towards}.  
Data efficient DRL for DPG-based dexterous manipulation has been further explored in \cite{popov2017data}, which mainly focuses on stacking Lego blocks.


\section{SYSTEM}\label{system}
\subsection{Preliminaries and notation}

Robotics Reinforcement Learning is a control problem in which a robot acts in a stochastic environment by sequentially  choosing  actions (e.g. torques to be sent to controllers) over  a  sequence  of  time  steps. The aim is that of  maximizing  a  cumulative  reward. Such problem is commonly modeled as a Markov Decision Process (MDP) that provides:  a state space $\mathcal{S}$, an action space $\mathcal{A}$, an initial state distribution with density $p(s_1)$, a stationary transition dynamics distribution with conditional density $p(s_{t+1}|s_t,a_t)$ satisfying the Markov property $p(s_{t+1}| s_1,a_1,\dots,s_T,a_T) $  for any  trajectory $\tau : \{s_1,a_1\dots,s_T,a_T\}$  and a reward function $r:\mathcal{S}\times\mathcal{A}\to \mathbb{R}$ .
The policy $\pi$ (i.e., the robot controller) mapping $\mathcal{S}\to\mathcal{A}$ is used to select actions in the MDP. The policy can be stochastic $\pi(a|s): s\mapsto \Pr[a|s]$ or deterministic $\mu: s \mapsto a=\mu(s)$. In DRL the policy is commonly parametrized as a DNN and denoted by $\pi_\theta$, where $\theta$ is the general parameter storing all the network's weights and biases. A typical example is a gaussian Multi-Layer Perceptron (MLP) net, which samples the action to be taken from a gaussian distribution of actions over states: \begin{equation}\label{gaussianMLP}
\pi_\theta(a|s) = \dfrac{1}{\sqrt{(2\pi)^{n_a} \det \Sigma_\theta(s)}}\exp\left( -\dfrac{1}{2}||a-\mu_\theta(s)||_{\Sigma_\theta^{-1}(s)}^2\right)
\end{equation}
The return $R_t^\gamma=\sum_{k=t}^{T}\gamma^{k-t}r_k$, with $r_t=r(s_t,a_t)$, is the total discounted reward from time-step $t$ onwards, where $\gamma\in]0,1[$ is known as a discount factor that favors proximal rewards instead of distant ones.  In RL value  functions  are  defined  as the  expected  total  discounted reward: state-value $V^{\pi}(s)= \mathbb{E}[R_1^\gamma|s,\pi]$ and action-value $Q^{\pi}(s,a)= \mathbb{E}[R_1^\gamma|s,a,\pi]$.
DRL methods usually approximate such value functions with neural networks (critics) and fit them empirically on return samples with stochastic gradient descent on a quadratic Temporal Difference (TD) loss.
The agent's goal is to obtain a policy which maximises the return from the initial state, denoted by the performance objective $J(\pi)=\mathbb{E}[R_1^\gamma|\pi]$. To do so, classical RL methods pick the best action that maximize such value functions (acting greedily) while sometime acting randomly to explore $\mathcal{S}$. This fact is taken into account in DRL with stochastic policies or with deterministic policies with added noise. Since not every robotic setup may have the possibility to inject noise into the controller for space exploration, we explored both stochastic and deterministic model-free DRL algorithms. In this paper, we implemented a Trust Region Policy Optimization (TRPO) \cite{schulman2015trust} as a stochastic policy and a Deep Q-Network with Normalized Advantage Functions (DQN-NAF) \cite{gu2017deep} as a deterministic one. 

\subsection{Algorithms}
Policy gradient (PG) methods are a class of RL algorithms that enjoy many good convergence properties and model-free formulations. The main reason that led to PG methods is that the greedy update of classical RL often leads to big changes in the policy, while in a stable learning it is desirable that both policy and value function evolve smoothly. Thus it is preferable to take little steps in the parameter space ensuring that the new policy will collect more reward than the previous one. The direction of the update should be provided by some policy gradient, which must be estimated as precise as possible to secure stable learning. 
The general stochastic gradient ascent update rule for PG methods is
\begin{equation}
\theta\gets \theta+\alpha\nabla_\theta J(\theta)
\end{equation}
where $\alpha$ is the learning rate. A proper network optimizer with adaptive learning rate such as Adam \cite{kingma2014adam} is strongly advised for such updates. Vanilla Policy Gradient (VPG), a variant of REINFORCE algorithm \cite{williams1992simple}, estimates the policy gradient from $N_\tau$ policy rollouts with the log-likelyhood ratio trick formula:
\begin{equation}
    \widehat{\nabla_\theta J}(\theta)=\frac{1}{N_\tau}\sum_{i=1}^{N_\tau}\sum_{t=1}^{T}\nabla_\theta\log\pi_\theta(a_{t,i}|s_{t,i}) R_t^\gamma
\end{equation}
	where $R_t^\gamma$ is a single sample estimate of $Q^{\pi_\theta}(s,a)$, thus typically with high variance. Many methods have been proposed to reduce the PG variance, including another neural net for estimating $Q^\pi$ or $V^\pi$ (actor-critic methods), or the use of importance sampling to reuse trajectories from older policies. In this paper we use TRPO and DQN-NAF. 
	
	TRPO \cite{schulman2015trust,achiam2017constrained} can be seen as ab advanced version of PG (Algorithm 1). Three are the main improvements:
	\begin{itemize}
	    \item[1)] $R_t^\gamma$ is replaced with lower variance \textit{advantages} $A^\pi(s,a)= Q^\pi(s,a)-V^\pi(s)$, estimated with Generalized Advantage Estimation algorithm ($\lambda$)\cite{schulman2015high} (similar to actor-critic algorithm).
	    \item[2)] it uses the Natural Policy Gradient (NPG), making the PG invariant to the parametrization $\theta$ used by premultiplying it with the inverse of the policy Fisher Info Matrix, namely the \textit{metric tensor} for policy space.
	    \begin{equation}
	        H_k=\mathbb{E}_{\tau\sim\pi_{\theta_k}}\left[ \nabla_\theta \log \pi_\theta(a|s)\nabla_\theta \log \pi_\theta(a|s)^{\top}\right]\Big|_{\theta=\theta_k} 
        \end{equation}
    This kind of update takes into account also the distance in KL-divergence terms between subsequent policies. Bounding such divergence helps in stabilizing the learning. Finally, since for neural network policies with tens of thousands of parameters NPG incurs prohibitive computation cost by forming and inverting the empirical FIM. Therefore is it usually retrieved approximately using a Conjugate Gradient (CG) algorithm with a fixed number of iterations.
    \item[3)] A line search algorithm is performed to check if there has been an improvement in the \textit{surrogate} loss $\mathcal{L}_\pi(\pi')=J(\pi')-J(\pi)\geq0$ and the old policy does not differ too much from the updated one in distribution.
	\end{itemize}
	This algorithm proved very successful in contacts rich environment and high-dimensional robots for locomotion tasks, but its efficiency in common robotic tasks such as 3D end-effector positioning and Pick\&Place must be yet validated.

\begin{algorithm}[t!]
	\caption{Trust Region Policy Optimization (TRPO) }
	\begin{algorithmic}[1]
		\State Randomly initialize policy parameters $\theta_0$
		\For{$k=0,1,2,\dots,$}
		\State Collect set of $N_\tau$ trajectories under policy $\pi_{\theta_k}$
		\State Estimate advantages with GAE($\lambda$) and fit $V^{\pi_{\theta_k}}$
		\State Estimate policy gradient	$$\hat{g}_k=\dfrac{1}{N_\tau}\sum_{i=1}^{N_\tau} \sum_{t=0}^{T-1}\gamma^{t} \nabla_\theta \log \pi_\theta (a_{t,i}|s_{t,i})\big|_{\theta=\theta_k}\hat{A}(s_{t,i},a_{t,i})$$
		\State Estimate $\hat{H}_k=\nabla^2_\theta D_{KL}(\pi_\theta||\pi_{\theta_k})\Big|_{\theta=\theta_k}$
		\State Compute $\hat{H}_k^{-1}\hat{g}_k$ with CG algorithm 
		\State Compute policy step $\Delta_k=\sqrt{\frac{2\delta_D}{\hat{g}_k^\top \hat{H}^{-1}_k\hat{g}_k}}\hat{H}_k^{-1}\hat{g}_k$
		    \For{$l=1,2,\dots,L$}
		    \State Compute candidate update $\theta^c=\theta_k+\nu^l\Delta_k$
		        \If{ $\mathcal{L}_{\pi_{\theta_k}}(\pi_{\theta^c})\geq0 \text{ and } D_{KL}(\pi_{\theta^c}||\pi_{\theta_k})\leq \delta_D$}
		            \State Accept candidate $\theta_{k+1}=\theta^c$
		        \EndIf
		    \EndFor
		\EndFor
	\end{algorithmic}\label{TRPO}
\end{algorithm}

DQN-NAF was proposed by \cite{gu2016continuous} and aims to extend Q-learning to continuous spaces without relying on PG estimates. Therefore, in order to solve the hard problem of $Q$-maximization in continuous action space, \cite{gu2016continuous} proposed the introduction of Normalized Advantage Functions (NAF).
 This new method, which adopts a deterministic neural network policy $\mu(s)$, enforces the advantage function to be shaped as a second order quadratic convex function, such as
\begin{equation}\label{NAFadvantage}
A^{\mu}(s,a)= -\dfrac{1}{2}||a-\mu(s)||^2_{P(s)}
\end{equation}
where $P(s)$ is a trainable state-dependent positive definite matrix.
Since $V^{\mu}(s)$ acts just as a constant in the action domain and that $Q^\mu(s,a)=A^\mu(s,a)+V^\mu(s)$, the final $Q$-function has the same quadratic properties of the advantage function (\ref{NAFadvantage}) and it can be easily maximized by choosing always $a=\mu(s)$.
This allows to construct just one net with that will output $P(s), V(s)$ and $\mu(s)$ to retrieve the $Q$-values. Clearly the overall $Q$-network parameter $\theta^Q$ is the union of $\theta^V,\theta^\mu$ and $\theta^P$, since they differ only in the output layer connection. The DQN-NAF pseudocode is presented in Algorithm \ref{DQN-NAF}.
The structure is very similar to DDPG due to use of targets nets for computing the TD loss but uses only one more complex $Q$-network that incorporates the policy. Another slight difference is that the critic may be fitted $K_Q$ times each timestep, acting as a critic-per-actor update ratio. This increases computational burden but stabilizes even more learning since the state-value network approximates better of the true $V^\mu(s)$, improving policy updates reliability.
This algorithm was applied with success directly onto a 7 DOF robotic arm in \cite{gu2017deep}, even managing to learn how to open a door from scratch. In particular it was implemented an asynchronous version of DQN-NAF surfing the ideas of \cite{mnih2016asynchronous}, where multiple agents were collecting samples to be sent to a shared replay buffer. In this way learning is almost linearly accelerated with the number of learners, since the replay buffer $\bm{\mathscr{R}}$ provides more decorraleted samples for the critic update. Obviously the reward function plays an important role in both DDPG and DQN-NAF and we will focus on different designs to explore the performances on these two state-of-the-art DRL algorithms.
\begin{algorithm}[t!]
    \caption{Deep Q-Network with NAF (DQN-NAF)}
    \begin{algorithmic}[1]
	    \State Randomly initialize $Q$ and target $Q'$ with $\theta^{Q'}\gets \theta^Q$
	    \State Allocate Replay buffer $\bm{\mathscr{R}}$
	    \For{episode $1,\dots,N_\tau$}
	        \For{$t=1,\dots,T$}
	            \State Execute $a_t=\mu_{\theta^\mu}(s_t)$
	            \State Store in $\bm{\mathscr{R}}$ transition $(s_t,a_t,r_t,s_{t+1})$
	            \For {iteration $k=1,\dots,K_Q$}
	                \State Sample minibatch of $N_b$ transitions from $\bm{\mathscr{R}}$ 
	                \State Set targets $y_i=r_i+\gamma V_{\theta^{V'}}(s_{t+1})$
	                \State Update $\theta^Q$ by minimizing loss $$L(\theta^Q)=\frac{1}{N_b}\sum_{i=1}^{N_b}\Big(y_i-Q_{\theta^Q}(s_i,a_i)\Big)^2$$
	                \State Update target network $$\theta^{Q'}\gets \xi\theta^{Q} +(1-\xi)\theta^{Q'}$$
	            \EndFor
	        \EndFor
	    \EndFor
    \end{algorithmic}
    \label{DQN-NAF}
\end{algorithm}


\section{SIMULATED EXPERIMENTS}\label{simulated_experiments}
We first compared the most promising state of the art algorithms by means of simulated tasks modeled using the MuJoCo physics simulator~\cite{todorov2012mujoco}. Simulation lets fast and safe comparisons of design choices such as, for DRL, the hyperparameters' setting. We modeled a UR5\footnote{\href{https://www.universal-robots.com/products/ur5-robot/}{https://www.universal-robots.com/products/ur5-robot/}} manipulator robot from Universal Robots with a Robotiq S Model Adaptive 3-fingers gripper\footnote{\href{https://robotiq.com/products/3-finger-adaptive-robot-gripper}{https://robotiq.com/products/3-finger-adaptive-robot-gripper}} attached on its end effector, for a total of 10 degrees of freedom. The same robot was used in our real-world experiments. We want to emphasize the fact that only one robotic arm was modeled for the simulated experiments in order to keep consistency with the real-world setup. However, analyzed algorithms would remain robust even in case of changes of the dynamic and geometric models of the robot (e.g., link lenghts, masses, and inertia).

\subsection{Robot Modeling}

The manipulator and gripper MuJoCo models (MJCF files) are generated from the robots' Unified Robot Description Formats (URDFs)\footnote{\href{http://wiki.ros.org/urdf}{http://wiki.ros.org/urdf}}. Once attached the MJCF files to each other, we computed the following global joint state and torque

\begin{equation}
q=\begin{bmatrix}
q_{\text{ur5}}\\
q_{\text{grip}}
\end{bmatrix}\in \mathbb{R}^{18}\qquad a=\begin{bmatrix}
a_{\text{ur5}}\\a_{\text{grip}}
\end{bmatrix}\in \mathbb{R}^{10}
\end{equation}

where $q_{\text{ur5}}=\begin{bmatrix}q_1&q_2&q_3&q_4&q_5&q_6\end{bmatrix}^\top$ and $q_{\text{grip}}=\begin{bmatrix}q_{10}&\dots&q_{13}&q_{20}&\dots&q_{23}&q_{30}&\dots&q_{33}\end{bmatrix}^\top$ ($q_{ij} | i = \{1, 2, 3\}$ is the $i$-th finger and $j=\{1, 2, 3\}$ is the $j$-th phalange)
are the UR5 and gripper joint positions vectors, respectively (measures expressed in radiants). $a_{\text{ur5}}=\begin{bmatrix}m_1&m_2&m_3&m_4&m_5&m_6\end{bmatrix}^\top$ and $a_{{grip}}=\begin{bmatrix}m_{00}&m_{11}&m_{12}&m_{13}\end{bmatrix}^\top$ are the UR5 and gripper \textit{action} vectors, i.e., the torques $m$ applied to each joint by its motor.


\begin{table}[t!]\centering
	\begin{tabular}{c c c c }\hline
		\multicolumn{1}{c}{Joint} & Joint Limits[\si{\radian}]& $k_r$	& $m_{\text{nom}}$ [\si{\newton\metre}] \\ \hline
		$q_1$ &[$-\pi,\pi$] & 101 & 150 \\
		$q_2$ &[$-\pi,0$]& 101 & 150 \\
		$q_3$ &[$-\pi,\pi$]& 101 & 150 \\
		$q_4$ &[$-\pi,\pi$]& 101 & 28\\
		$q_5$ &[$-\pi,\pi$]&101 &28\\
		$q_6$ &[$-\pi,\pi$]&101 &28\\
		\hline
	\end{tabular} 
\caption{UR5 Motor and Joint specifications.}\label{motorspecs}
\end{table}


\begin{table}[t!]\centering
	\begin{tabular}{c c c c }\hline
		\multicolumn{1}{c}{Joint} & Joint Limits [\si{\radian}]& $k_r$	& $m_{\text{nom}}$ [\si{\newton\metre}] \\ \hline
		$q_{i0}$ &$[-0.2967, 0.2967]$ & 14 & 0.8 \\
		$q_{i1}$ &$[0, 1.2217]$& 14 & 0.8 \\
		$q_{i2}$ &$[0, \pi/2]$& - &  -\\
		$q_{i3}$ &$[-0.6632, 1.0471]$& - & -\\
		\hline
	\end{tabular} 
	\caption{Robotiq 3-Finger Gripper Motor and Joint specifications.}\label{gripmotorspecs}
\end{table}

In order to better match the real robot, the MuJoCo model includes the actual gear reduction ratios $k_r$ and motors nominal torques $m_{\text{nom}}$ of Tab \ref{motorspecs} and \ref{gripmotorspecs}. Actuators were modeled as torque-controlled motors. As advised by MuJoCo documentation, joint damping coefficients were added and chosen by trial and error, resulting in an improved simulated joint stiffness.

Focusing on the gripper, its fingers under-actuated system was modeled as a constraint of joint phalanges angles. This joint coupling was implemented by defining fixed tendons lengths between phalanges through a set of multiplicative joint coefficients $c$. These parameters were found by trial and error until a satisfying grasp was obtained: $c_{12}=-1.5$ for the tendon between $q_{i1}$ and $q_{i2}$, $c_{23}=-3.5$ between $q_{i2}$ and $q_{i3}$, $\forall i=1,2,3$.
This is not how the real system works, but it is the best demonstrated way to ensure a correct simulated gripper closure. Finally, inertia matrices were correctly generated through the MuJoCo \texttt{inertiafromgeom} option, which enables automatic computation of inertia matrices and geoms frames directly from model's meshes.

An important parameter is the MuJoCo simulation timestep $T_M$, i.e., the timestep at which the MuJoCo Pro physics engine computes successive evolution states of the model, given an initial joints configuration. Usually, magnitude of milliseconds is chosen. In our case, $T_M=2 \si{\milli\second}$ ensures a good trade-off between simulation's stability and accuracy. Standard gravity ($9.81$ \si{\meter\second\squared}) was already enabled by the simulator by default.

In order to match the real UR5 controller, which operates the robotic arm at $f=125 \si{\hertz}$, we set a \texttt{frameskip} $N_{T_M}=4$. This value defines how many MuJoCo state evolutions the OpenAI's Gym environment must skip, with an effective sampling time $T_s$ of 
\begin{equation}
T_s= T_M\cdot N_{T_M} =8 \text{\si{\milli\second}}, \qquad f= \dfrac{1}{8 \text{ \si{\milli\second}}}=125 \text{ \si{\hertz}} 
\end{equation}
This method guarantees a stable and accurate simulation while sampling our modeled system at the correct rate. 

\subsection{Tasks}\label{tasks}

\subsubsection{Random Target Reaching}

\begin{figure}[t!]
	\centering
	\includegraphics[scale=0.2]{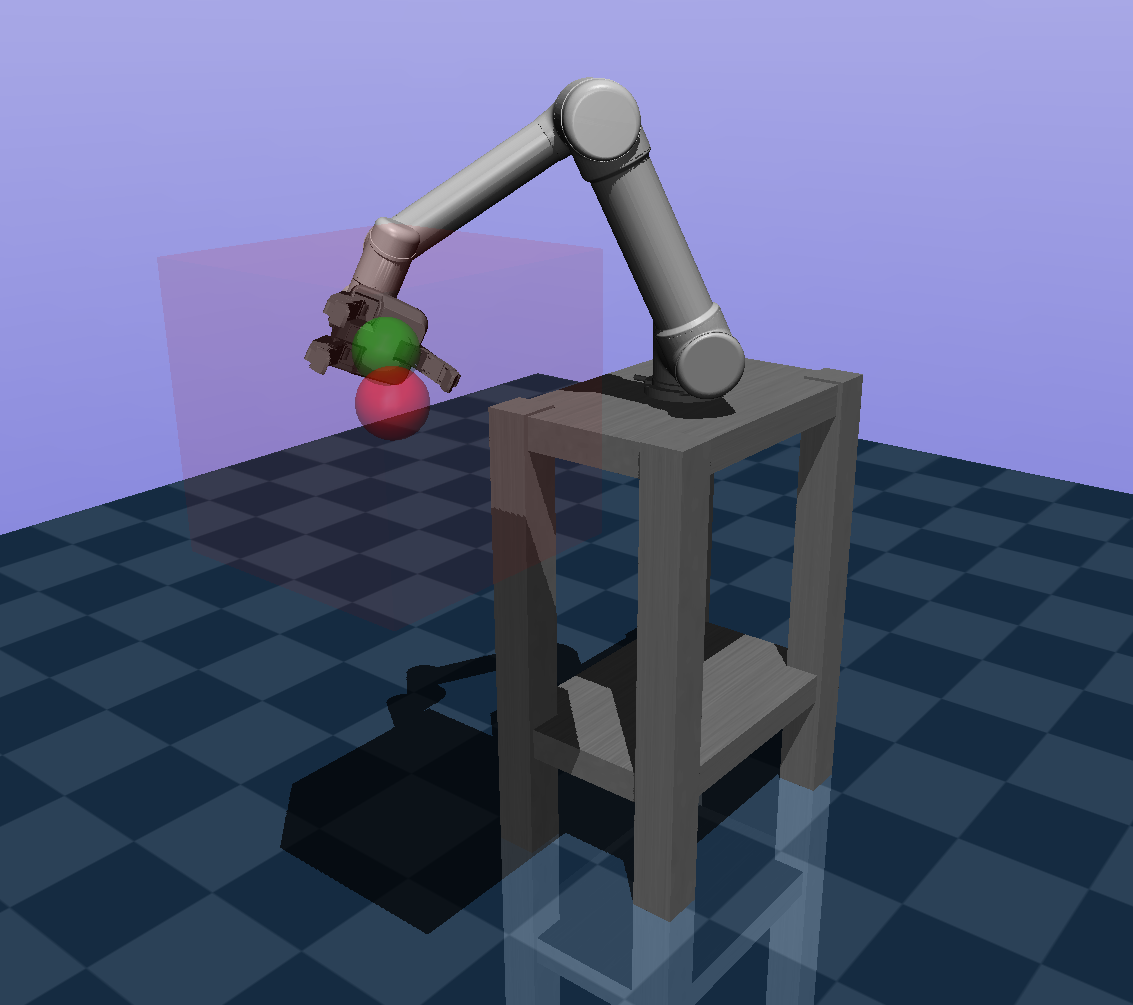}
	\caption{UR5 Reach Task MuJoCo Environment}\label{reach_task}
\end{figure}

The robot end effector must reach a random target position $p_\text{goal}$ (the center of the red sphere of Figure \ref{reach_task}) within a fixed cube of side 40 \si{\centi \meter} in the robot workspace. In global coordinates (world reference frame positioned and centered on the floor under the robotics arm bench): 
\begin{equation}\label{redbounds}
p_{\text{goal}}=U_3[-0.2,0.2]+\begin{bmatrix}
0.7&0&0.9
\end{bmatrix}^\top
\end{equation}
where $U_3[-B,B]$ is a $3\times1$ vector whose entries are sampled uniformly within the specified bounds $\pm B\in \mathbb{R}$.

The choice of restricting the goal position within a cube aims to limit the training space of DRL algorithms, otherwise extended 850 \si{\milli \meter} from the base joint of the robot. 
In order to promote space exploration and avoid deterministic behavior, uniformly sampled noise is added to the initial joint positions and velocities of the UR5:
\begin{equation}
\begin{aligned}
q_{\text{ur5}}&=q_{\text{ur5},0}+U_6[-0.02,0.02] \\
\quad q_{\text{ur5},0}&=\begin{bmatrix}-0.3&-0.7&1.1&-0.1&-0.2&0 \end{bmatrix}^\top \text{ \si{\radian}}\\
\dot{q}_{\text{ur5}}&=\dot{q}_{\text{ur5},0}+U_6[-0.1,0.1] \\
\quad\dot{q}_{\text{ur5},0}&= \begin{bmatrix}
0&\cdots&0
\end{bmatrix}^\top\text{ \si{\radian\per\second}}
\end{aligned}
\end{equation}
The state of the environment follows:
\begin{equation}
s=\begin{bmatrix}
q^\top&\dot{q}^\top&p_{\text{goal}}^\top&p_\text{ee}^\top
\end{bmatrix}^\top\in \mathbb{R}^{42}
\end{equation}
where $q$ is the robotic arm joint vector; $\dot{q}$ is its time derivative; $p_\text{ee}$ and $p_\text{goal}$ are the position of the end effector and of the target, respectively.

The episode horizon for this task has been set to $T=300$, which means that the agent is allowed to achieve its goal within
$T\cdot T_s= 2.4 \text{ \si{\second}}$. Thus the engine computes a total of $2.4$ s of simulated time; after that the episode is terminated and a new one starts.

The reward function follows: 
\begin{equation}
r_{\texttt{R}}(s,a)=-||p_{\text{goal}}-p_{\text{ee}}||-c_a||a||
\end{equation}
The regularization term $c_a||a||$ aims to promote the learning of stable and bounded actions, slightly penalizing ($c_a\approx 10^{-3}$) the usage of excessive torques.
This reward function is always negative (penalizing rewards) thus the maximum collectible return is $0$. Here we can define a particular environment state $s$ as terminal by checking if the task has been correctly performed truncating the episode. However in this case the agent must experience the whole trajectory until the episode horizon threshold $T$ if a previous good terminal state is not encountered. Any other type of termination will lead to higher return, tricking agent to infer the actual sequence of action as good. Due to this fact, such a reward slows the initial learning process since it is highly likely that the robots may find itself in a state far from optimum but still it must experience the whole bad episode.
A discrete timestep
\begin{equation}
t \text{ is terminal if } \quad ||p_\text{ee}(t)-p_\text{goal}(t)||<5\text{ \si{\centi\meter}}
\end{equation}
that means the assigned task is achieved.

\subsubsection{Pick\&Place}

\begin{figure}[t!]
	\centering
	\includegraphics[scale=0.2]{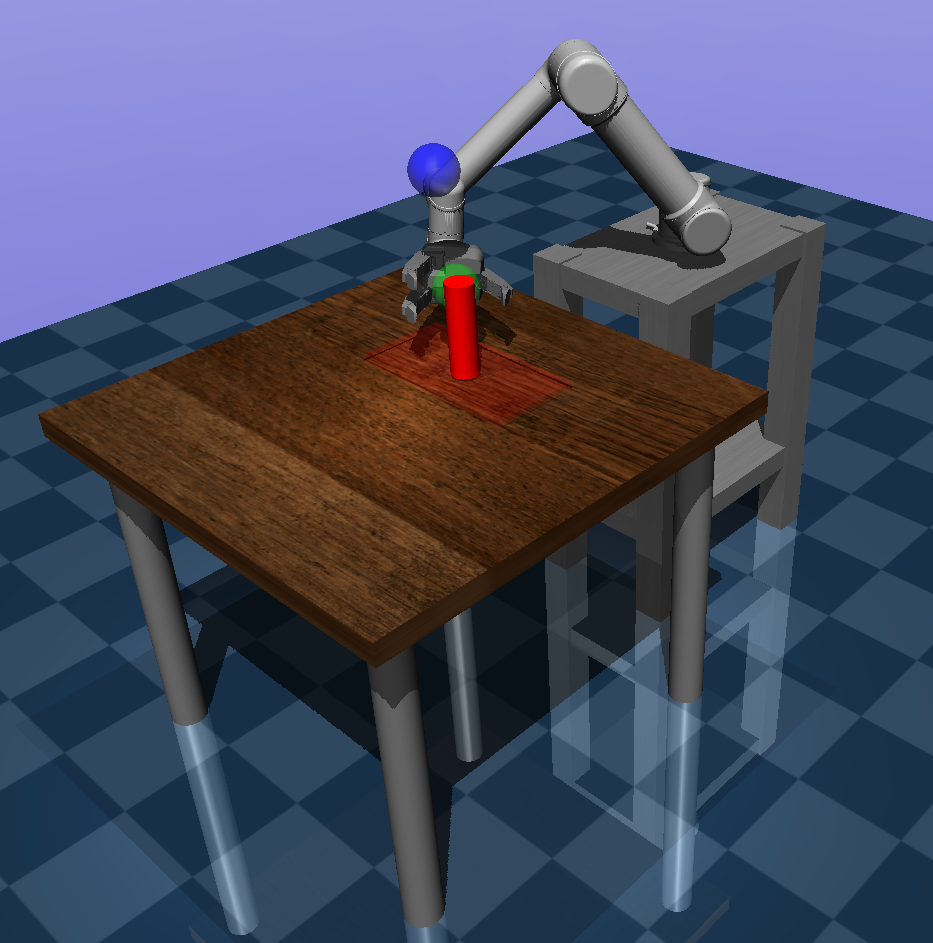}
	\caption{UR5 Pick\&Place Task MuJoCo Environment}\label{pick}
\end{figure}

The arm must learn how to grasp a cylinder from a table ($R_\text{cyl}=2$ \si{\centi\meter}, $mass_\text{cyl}=0.5$ \si{\kilogram}) and place it about $30\si{\centi\meter}$ above the object (see Figure \ref{pick}):
\begin{equation}
p_{\text{cyl}}= \begin{bmatrix}
0.7&0&0.9
\end{bmatrix}^\top\qquad p_\text{goal}= \begin{bmatrix}
0.7&0&1.2
\end{bmatrix}^\top
\end{equation}
At every episode, both cylinder and goal positions are fixed, while the initial position of the robot's joints is uniformly sampled.

The state of the environment follows:
\begin{equation}
s=\begin{bmatrix}
q^\top&
\dot{q}^\top&
p_\text{ee}^\top&
p_\text{goal}^\top&
p_\text{cyl}^\top
\end{bmatrix}^\top\in\mathbb{R}^{45}
\end{equation} 
$T=500$ is selected as timesteps, that means a total allowable time to perform the task equal to 
$T\cdot T_s= 4 \text{ \si{\second}}$. A similar task was already performed in \cite{gu2017deep} with DQN-NAF, but with a stick floating in the air attached to a string and a simplified gripper with fingers without phalanges. Our task instead is more realistic and the robot must learn to firmly grasp without any slip the cylinder. 

Inspired by \cite{gu2017deep}, we created a geometrical-based reward function that promotes the minimization of three distances:
\begin{itemize}
	\item the distance from the end effector to the object: $d_1=||p_\text{ee}-p_\text{cyl}||$
	\item the distance from the fingers to the center of mass of the cylinder: $$d_2=\sum_{i=1}^{3}\Big( ||p_\text{cyl}-p_{f_i}||-R_\text{cyl}\Big) $$
	In particular $p_{f_i}$ is the $3\times1$ cartesian position of the second phalanx of finger $i$ in the world reference frame. The radius of the cylinder $R_\text{cyl}$ acts simply as offset to avoid nonsense penalties since it is impossible to reach with the fingers the object's center of mass.
	\item the distance from the cylinder to the goal: $d_3=||p_\text{cyl}-p_\text{goal}||$
\end{itemize}
The final reward function is:
\begin{equation}
r_{\texttt{P}}(s,a)=\dfrac{1}{1+\sum_{j=1}^{3}c_jd_j}-c_a||a||
\end{equation}
where $c_j$ is manually selected in order to balance distance weightings. The function is normalized in order to avoid huge rewards when reaching the goal. In this way, when no torque is applied to motors and the goal is reached, the highest reward possible is 1.
This is an encouraging reward function: such reward shaping is one of our main novelties and it foresee that its values are instead mainly positives, allowing us to define a \textit{bad} terminal state and speeding up simulation of many trajectories. This type of reward function is widely diffused in locomotion tasks, since it is easy to assign a reward proportional to the distance traveled or forward velocity. On the other hand for robotic manipulations this is not always trivial and such a reward function can be hard to compose efficiently.

The gripper must stay close to the cylinder and the cylinder to the goal, that means the episode is terminated on the following state check:
\begin{equation}\label{terminalPv2}
t \text{ is terminal if } \quad\begin{cases}
d_1(t)> 0.35 \text{ \si{\meter}}\\
d_3(t)> 0.35\text{ \si{\meter}}\\
\end{cases}
\end{equation}

\subsection{Hyper-parameters settings}

By trial and error, we found that $N_\tau^\text{max}= 4000 $ episodes guarantees a good training for the reaching task while $N_\tau^\text{max}= 5000$ is a good trade-off for the Pick\&Place: the training is stopped when $N_\tau^{\text{max}}$ is reached.  \texttt{rllabplusplus} algorithms perform the policy update every $\mathcal{B}$ samples. This means that every algorithm iteration/policy update is done every $N_\tau=\mathcal{B}/T$ episodes, were $T$ is the maximum number of episode timesteps (max path length). Moreover, we used a discount factor $\gamma=0.99$ in order to make the agent slightly prefer near future rewards rather than distant ones.  Specifically for every algorithm:

\paragraph{DQN-NAF}
DQN-NAF updates the policy based on the critic estimation. The seamless integration of the policy in the second order approximated critic allows to select, at each timestemp, the action that globally maximize the $Q$ function. 
We tested three different minibatch sizes: $N_b=32,64,128$. In order to explore the fact that the same but scaled reward function may cripple the learning, only in the policy update we scaled the rewards by a factor $r_s=1,0.1,0.001$. In other words, the reward used to update the policy is 
\begin{equation}
r(s,a)= r(s,a)\cdot r_s
\end{equation}
 In principle a lower reward should reduce the base stepsize of the policy gradient. Intuitively this whole method is heavily task dependent but proved \cite{islam2017reproducibility} to stabilize (though slow down) the learning. The soft target update coefficient for target networks  used was left to the default value $\xi=10^{-3}$. 

\paragraph{TRPO}
We used the Conjugate Gradient (CG) Algorithm with $n_{CG} = 10$ iterations in order to estimate the NPG direction and to fit the baseline network. We used the \texttt{rllab} default trust region size $\delta=0.01$ for both policy ($\delta_D$) and baseline ($\delta_V$) updates. Tests demonstrated that the size of the baseline network does not significantly affect the learning progress; thus, it was fixed to $100\times100$. This might reflect the fact that the baseline is deep enough to effectively predict the states value it is fed with; a larger network would slow the training and introduce overfitting. The MLP baseline network is updated through the CG algorithm. For the advantage estimation procedure we used a  GAE coefficient $\lambda=0.97$ as suggested by \cite{schulman2015high}. According to \cite{islam2017reproducibility}, the batch size $\mathcal{B}$ highly affects the stability and the learning performance curve. Thus, we tested 3 different batch sizes, corresponding approximately to a $N_\tau=10,20,40$ environment runs per algorithm iteration.

\subsection{Evaluation and results}

The average return is used to evaluate policies performances. After each update of the $\pi_\theta$ policy neural network, the new controller is tested on $N_\tau^\text{test}=10$ new task episodes and an estimate of the agent performance $J(\pi_\theta)=\E_{\tau \sim\pi_\theta}\left[r(\tau)\right]=\E_{\pi_\theta}\left[\sum_{t=1}^{T}r(s_t,a_t)\right]$ is estimated, i.e., the average undiscounted return $\bar{r}$ along with its standard deviation $\sigma_r$:
\begin{equation}
\bar{r}=\dfrac{1}{N_\tau^\text{test}} \sum_{i=1}^{N_\tau^\text{test}}r(\tau_i)\qquad \hat{\sigma}_r=\sqrt{\dfrac{1}{N_\tau^\text{test}}\sum_{i=1}^{N_\tau^\text{test}}\Big(r(\tau_i)-\bar{r} \Big)^2 }
\end{equation}
$\sigma_r$ represents the shaded region around the mean return. We used the undiscounted return as evaluation metric because it lets an easier understanding of the mean sequence of rewards if compared with its $\gamma$-discounted version. 

Finally, \textit{Final Average Return} describes the average return of the last 10 policy runs. \textit{Episodes Required} indicates the minimum number of episodes required to reach a performance similar to a final policy characterized by \textit{Final Average Return}. 

These settings are used to compare DQN-NAF and TRPO for the proposed tasks with respect to two widely used state of the art DRL algorithms: Vanilla Policy Gradient (VPG)~\cite{} and Deep Deterministic Policy Gradient (DDPG)~\cite{}. Our aim is that of proving the robustness and adaptability of proposed approaches with respect to the proposed tasks. For a exhaustive comparison, we tested 4 different types of nets: $32\times32$, $100\times100$, $150\times 100 \times 50$ and $400\times 300$ (see Table \ref{policy_arch}). Policy networks are trained with $\tanh$, while $Q$ networks and $V$ baselines are equipped with Relu nonlinear activation function.

\begin{table}[t!]
	\centering
	\label{my-label}
	\begin{tabular}{cccc}
		\hline
		\multicolumn{1}{c}{\multirow{2}{*}{Policy $\pi_{\theta}$}} & \multicolumn{1}{c}{\multirow{2}{*}{Policy Hidden Sizes}} & \multicolumn{2}{c}{$\dim(\theta)$}                                                                   \\ \cline{3-4} 
		\multicolumn{1}{c}{}                     & \multicolumn{1}{c}{}                      & \multicolumn{1}{c}{Reach} & \multicolumn{1}{c}{Pick\&Place} \\ \hline
		1  &$32\times32$            & 2762  &  2858                   \\
		2  &$100\times100$ 			& 15410  & 15710                       \\
		3  &$150\times100\times50$ 	& 757110  & 757560                         \\
		4  &$400\times300$          &  140510 &  141710                \\
		\hline
	\end{tabular}	
	\caption{Policy architectures tested on the three different environments and algorithms}
\end{table}\label{policy_arch}

\subsubsection{Random Target Reaching}

\begin{figure}[t!]
	\centering
	\includegraphics[scale=0.42]{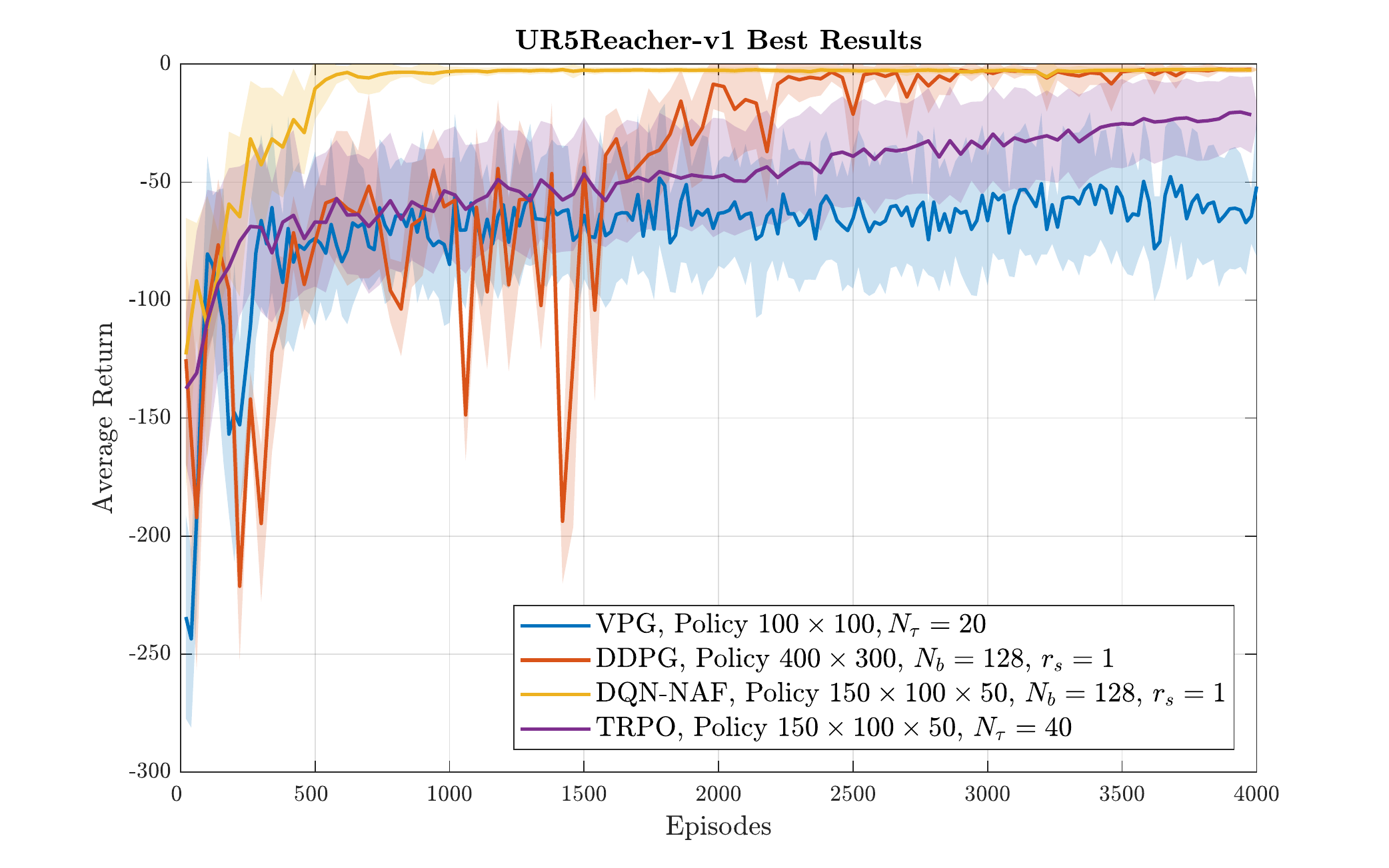}
	\caption{Random Target Reaching - Best results}\label{UR5Reacherv1FINAL}
\end{figure}

\begin{table}[t!]\centering
	\begin{tabular}{c c c c }\hline
		\multicolumn{1}{c}{Algorithm} & Episodes Req. & Final Avg Return & Max Return \\ \hline
		VPG & $260$ & $-60.90\pm24.70$ & $-47.65 \pm 24.71 $ \\
		DDPG &$2860$& $	-2.83 \pm  2.57$ & $ 	-2.13\pm1.21$ \\
		DQN-NAF &$700$& $-2.36\pm	1.23 $ & $	-2.18 \pm1.04$ \\
		TRPO &$2360$& $-22.87\pm16.27 $ & $-20.27\pm	14.83 $\\
		\hline
	\end{tabular} 
	\caption{Random Target Reaching - Numerical Results}\label{inforv1}
\end{table}

VPG struggles to learn a near-optimal policy (see Figure \ref{UR5Reacherv1FINAL}). The best VPG policy ($100\times100$) gets stuck after just $500$ episodes on an average return of about $-60$. TRPO is not able to solve the task but, thanks to its theoretical monotonic guarantees, it should be able to reach a close to zero return with a slightly higher number of episodes. DDPG can synthesize a $400\times300$ policy that achieves the best possible return in about $2500$ episodes. However, it is the algorithm with the most unstable return trend and it must be carefully tuned in order to get good results.

Being designed to perform robotic tasks, DQN-NAF stably solves the environment in less then $700$ episodes. Moreover, almost every policy architecture succeeds to collect almost zero return with a very similar number of episodes. This behavior uncorrelates the need for a huge net to perform the same task: it seems that it is the method the network is trained with that really makes the difference. However we cannot skip to test different nets on the next environments since this fact is surely related to the reward function used and the particular task. As a general rule, we found out that a net larger then $32\times32$ usually delivers better performance across these 4 algorithms.

\subsubsection{Pick\&Place} 
\begin{figure}[t!]
	\centering
	\includegraphics[scale=0.42]{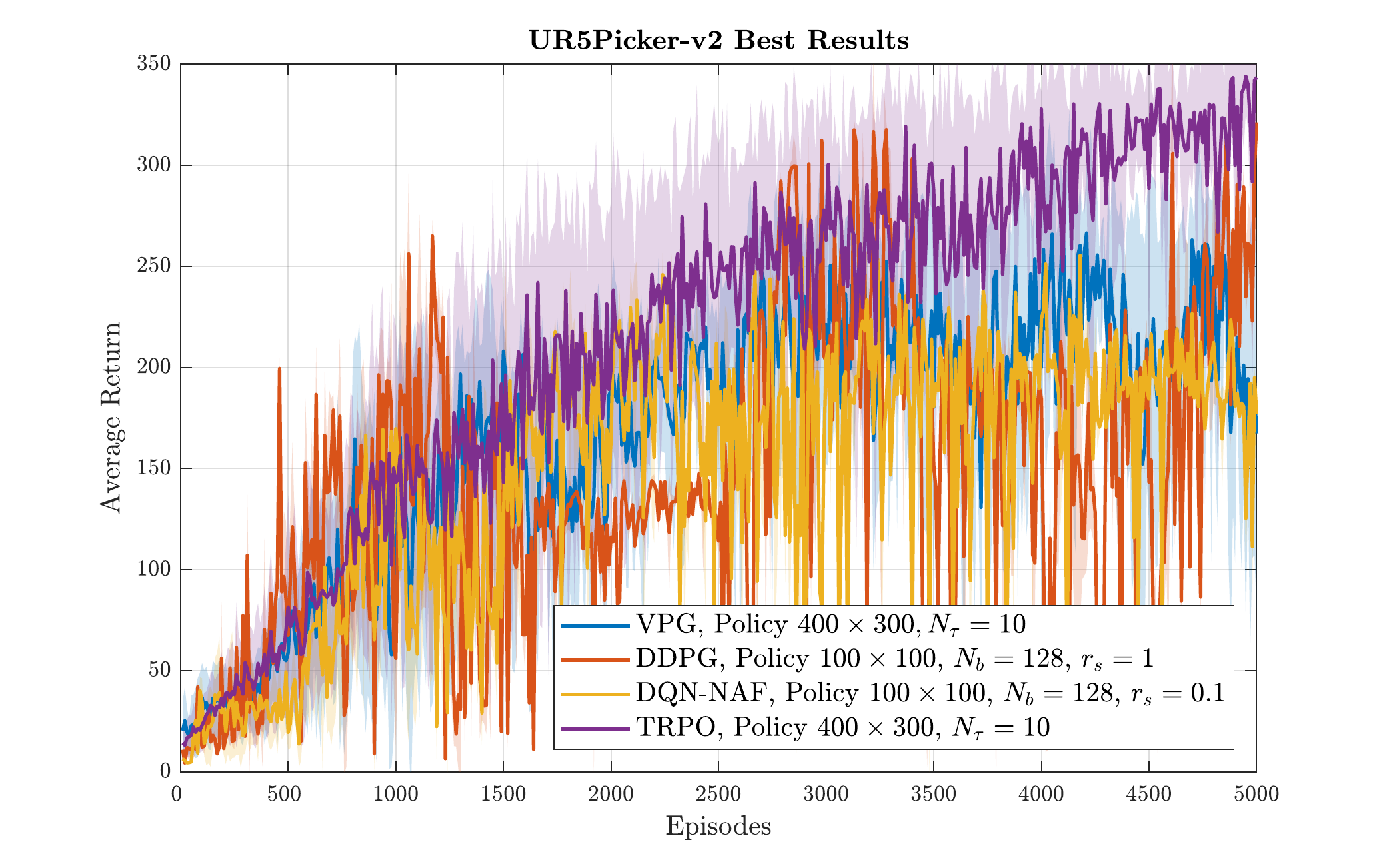}
	\caption{Pick\&Place - Best results}\label{UR5Pickerv2FINAL}
\end{figure}

As displayed in Figure \ref{UR5Pickerv2FINAL}, the pick and place environment proved highly stochastic due to the contacts between the gripper and the cylinder; little impacts during the grasp learning often lead the cylinder to fall and roll, preventing further grasp trials. This fact is reflected by the high average return variance and unstable learning in VPG, DDPG and DQN-NAF, for almost all network configurations. Their learning curves prove an overall return increase but the grasp still fails frequently due to the slipness of the cylinder or high approaching speed. The monotonic improvement theory and precautions of TRPO delivers after 5000 episodes an average return of $324\pm43$, performing a solid grasp while generating a stable trajectory for the cylinder placement on the blue goal. The most interesting fact about the TRPO grasp is the tilting of the cylinder towards the fingers. This allows the robotic arm to lift the cylinder with less effort while minimizing the risk of object slip/loss. On the other hand, the overall movements for the cylinder's transport can be sometimes more shaky than those observed in the reaching task with DQN-NAF. TRPO's policy was also chosen to perform the task on the real setup because it had the most room for improvement and further training may polish the network's behavior or deliver better grasping results.

\color{red}

\color{black}

\begin{table}[t!]\centering
	\begin{tabular}{c c c c }\hline
		\multicolumn{1}{c}{Algorithm} & Episodes Req. & Final Avg Return & Max Return \\ \hline
		VPG & $740$ & $187.11\pm70.63$ & $273.95 \pm 18,21 $ \\
		DDPG &$4950$& $	257.02 \pm  12.59$ & $321.24\pm14.37$ \\
		DQN-NAF &$1980$& $	173.08\pm	9.38 $ & $	255.39 \pm5.73$ \\
		TRPO &$4980$& $324.97\pm43.35 $ & $344.01\pm17.74 $\\
		\hline
	\end{tabular} 
	\caption{Pick\&Place - Numerical Results}\label{infopv2}
\end{table}

\section{REAL-WORLD EXPERIMENTS}\label{real_experiments}
Real-world experiments aim to prove that the policies learned in simulation are powerful also in real environments.

In order to use the learned policies in a real environment, it is necessary to put in communication the real setup with the simulated one.
The simulated environment can interface the external software by exchanging JSON data through a TCP Socket connection. As the real robotics setup is based on ROS, we used ROSBrige\footnote{\href{http://wiki.ros.org/rosbridge\_suite}{http://wiki.ros.org/rosbridge\_suite}} which provides a JSON API to ROS functionality for non-ROS programs.

Focusing on visual data, in order to easily obtain objects poses, fiducial markers are placed on them. In particular, we used the AprilTags~\cite{olson2011apriltag} library. 
A Microsoft Kinect One, placed in front of the robot, is used to view the scene.

\subsubsection{Random Target Reaching}
\begin{figure}[t!]
	\centering
	\includegraphics[width=0.4\textwidth]{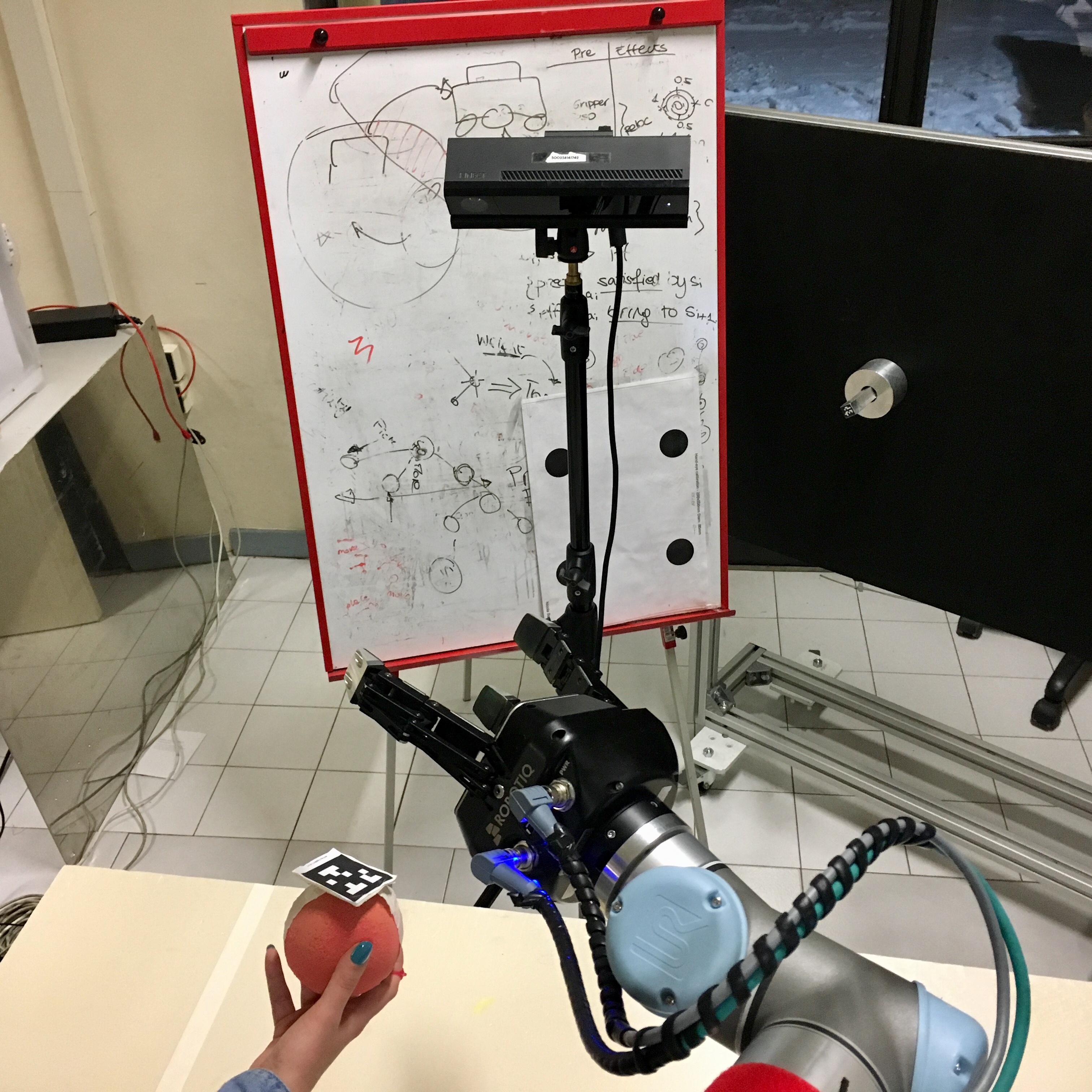}
	\caption{The Random Target Reaching experiment. The robot has to reach the red ball.}\label{UR5ReacherReal}
\end{figure}
The policy described in Section~\ref{simulated_experiments} was tested: a ball is sustained near the gripper as in Figure~\ref{UR5ReacherReal}. A marker is placed on it in order to obtain its pose. The robot is able to place its end effector at the ball position with a 100\% success rate. Moreover, the robot is able to follow the ball when in motion (see the supplementary video).

\subsubsection{Pick\&Place}
\begin{figure}[t!]
	\centering
	\includegraphics[width=0.4\textwidth]{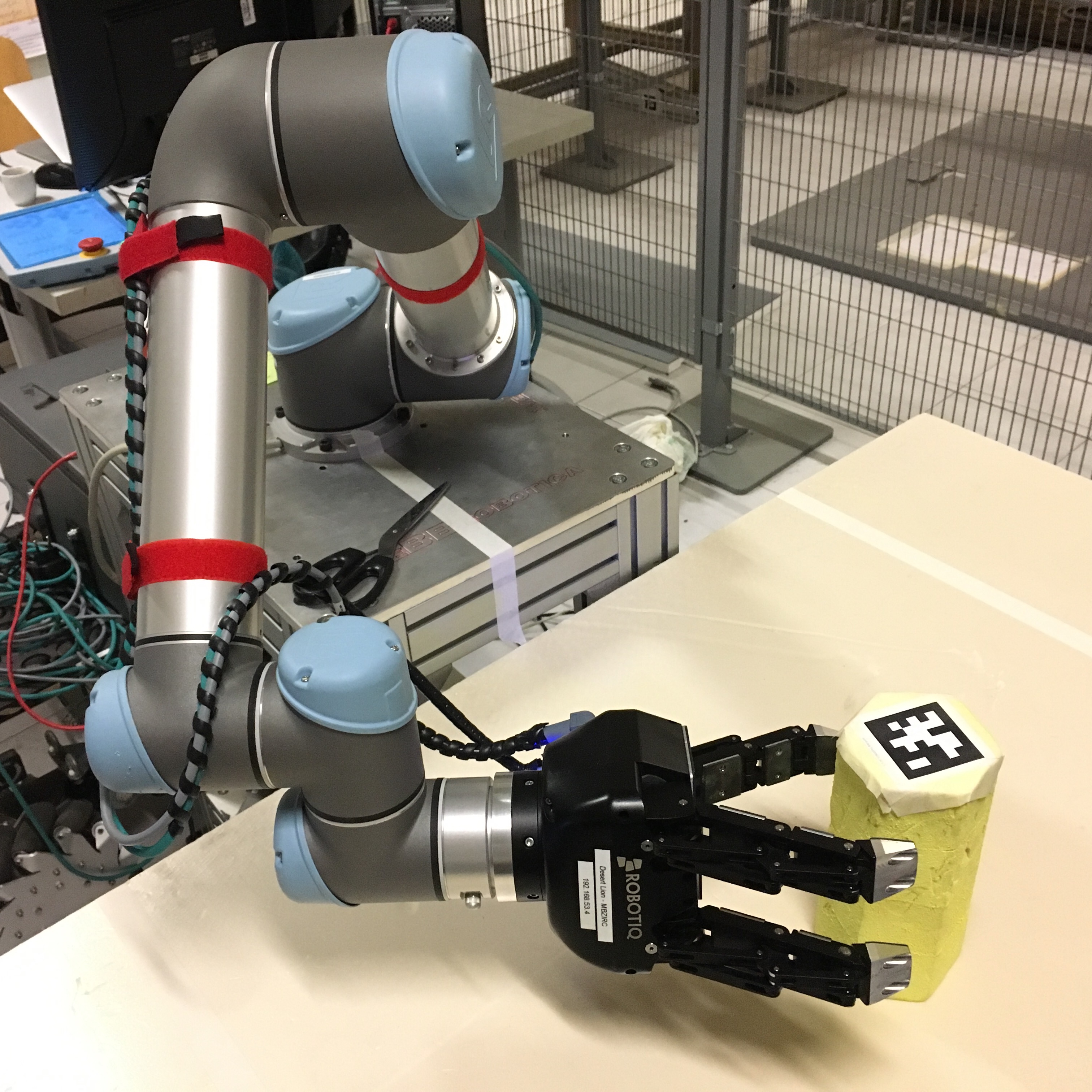}
	\caption{The Pick\&Place experiment. The robot has to pick up the yellow cylinder and bring it in a random place.}\label{UR5PickerReal}
\end{figure}
The robot has to pick up a cylinder placed on a table and bring it on a random point placed over the first one. As for the previous experiments, the cylinder pose is recognized using a fiducial marker (see Figure ~\ref{UR5PickerReal}). 100\% success is guarantees as demonstrated by the supplementary video.


\section{CONCLUSIONS AND FUTURE WORK}\label{conclusions}
Deep Reinforcement Learning algorithms provides nowadays very general methods with little tuning requirements, enabling tabula-rasa learning of complex robotic tasks with deep neural networks. Such algorithms showed great potential in synthesizing neural nets capable of performing the learned task while being robust to physical parameters and environment changes.  
In simulation, we compared DQN-NAF and TRPO 
to VPG and DDPG for classical tasks such as end-effector dexterity and Pick\&Place on a 10 DOF collaborative robotic arm. Simulated results proved that good performances can be obtained with reasonable amount of episodes, and training times can be easily improved with more CPUs on computational clusters. 
DQN-NAF performed really well on the reaching task, achieving a suboptimal policy. TRPO demonstrated to be the most versatile algorithm thanks to its reward scaling and parametrization invariances. VPG learns typically slower whereas DDPG is the most unstable and difficult to tune since it is highly reward scale sensitive. We discovered that the policy network architecture (width/depth) was not a decisive learning parameter and it is algorithm dependent. However, a hidden layer size of at least $100\times 100$ is advised for similar continuous control tasks. 
Finally we showed that it is possible to transfer the learned policies to real hardware with almost no changes.


\bibliographystyle{IEEEtran}
\bibliography{references}

\end{document}